\documentclass{article}

 \usepackage[preprint]{neurips_2026}

\usepackage[utf8]{inputenc} 
\usepackage[T1]{fontenc}    
\usepackage{hyperref}       
\usepackage{url}            
\usepackage{booktabs}       
\usepackage{amsfonts}       
\usepackage{nicefrac}       
\usepackage{microtype}      
\usepackage{xcolor}         
\usepackage{graphicx}
\usepackage{subfigure}
\usepackage{amsmath}
\usepackage{multirow}
\usepackage{multicol}
\usepackage{array}
\usepackage{makecell}

\usepackage{amssymb}
\usepackage{booktabs}
\usepackage{multirow}
\usepackage{bm}
\usepackage{bbding}
\usepackage{colortbl}
\usepackage{tabularx}

\usepackage{wrapfig} 
\usepackage{xcolor} 

\newcommand{\ours}{DVD}

\definecolor{citecolor}{HTML}{0071bc}

\usepackage{authblk}

\usepackage{soul}

\title{DVD: Dynamic Vector Decoding for Efficient MLLM-based Perception}

\author[ ]{\textbf{Jinghua Hou}}
\author[ ]{\textbf{Zhe Liu}\thanks{Project leader.}}
\author[ ]{\textbf{Hengshuang Zhao}\thanks{Corresponding author.}}
\affil[ ]{The University of Hong Kong} 

\affil[ ]{\textcolor{magenta}
{\url{https://almoonysl.github.io/projects/DVD}}}

\begin{document}

\maketitle

\begin{figure*}[h!]
\centering
\includegraphics[width=0.99\linewidth]{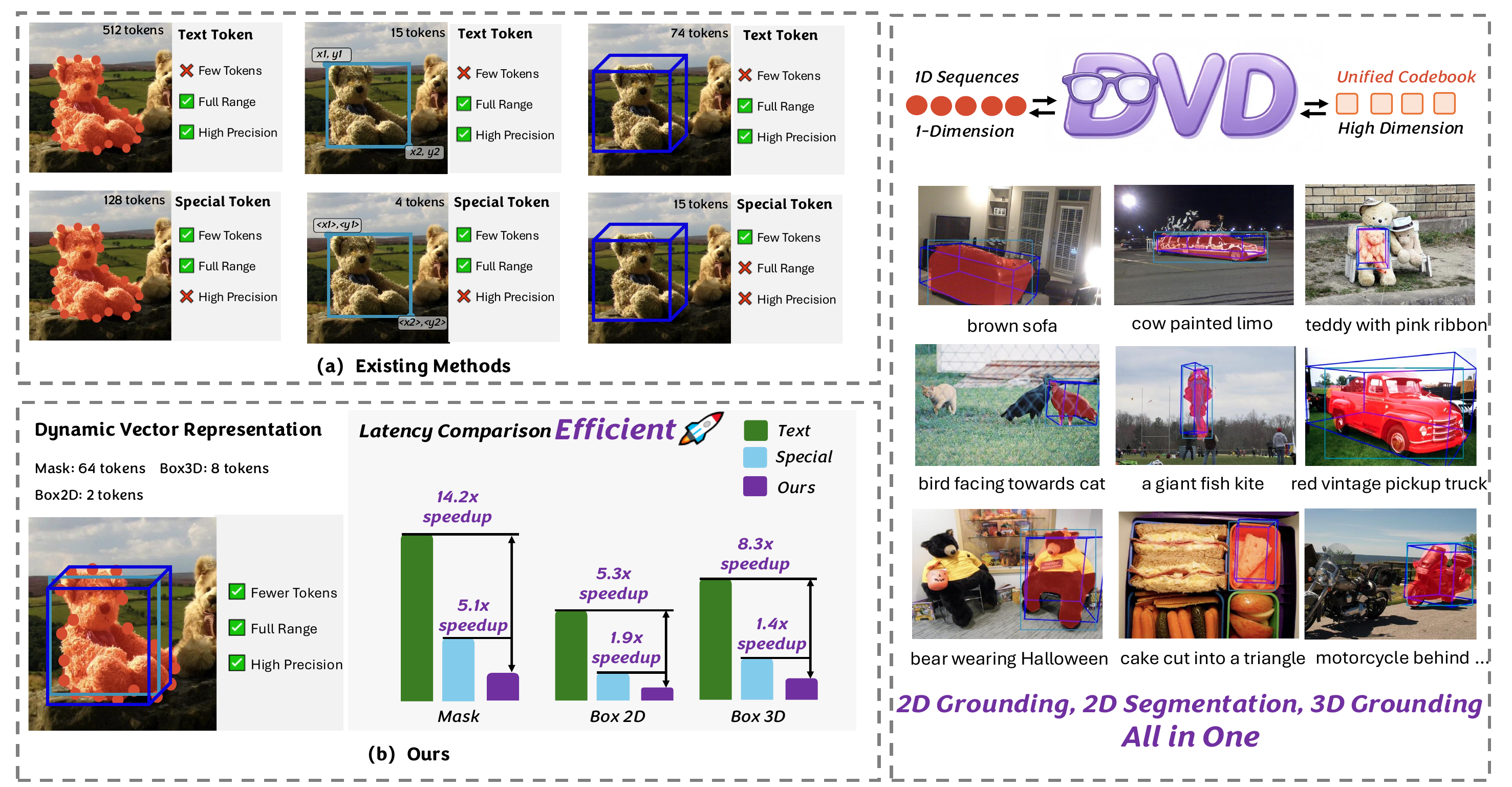}
\vspace{-5pt}
\caption{
Comparison of MLLM perception paradigms. Existing methods either use text tokens (high token count, e.g., 512 tokens for masks) or special coordinate tokens (limited range and precision) to encode perception outputs. Our dynamic vector representation unifies diverse perception representations by converting them into 1D sequences and learning a unified codebook in high-dimensional space. This yields compact and high precision representations with fewer tokens, delivering consistent latency reduction across 2D and 3D tasks.
}
\label{fig_intro}
\end{figure*}

\begin{abstract}
Multimodal large language models have made remarkable progress in bridging vision and language, facilitating various perception tasks essential for human-machine interaction, robotics, and autonomous driving. However, existing MLLM-based perception methods predominantly rely on text-based coordinate representation, which suffers from excessive token overhead, or fixed-range quantization, which suffers from range and precision constraints, especially for 3D domains with unbounded spatial range and high localization accuracy requirements. To address these challenges, we propose a dynamic vector decoding method named DVD, which unifies the representation of 2D and 3D perception tasks. Specifically, we first transform diverse perceptual representation (i.e., 2D bounding boxes, 2D masks, and 3D bounding boxes) into 1D vector sequences, which are then mapped to compact discrete tokens in the high-dimensional space. Then, a lightweight de-tokenizer enables seamless integration with MLLMs by decoding output tokens back to original 2D and 3D perceptual representations. Extensive experiments on 2D and 3D perception benchmarks including RefCOCO series, SUN-RGBD, KITTI, Hypersim, nuScenes demonstrate that DVD achieves superior performance in 2D and 3D tasks and reduces significantly the token overhead and inference latency. DVD provides an efficient and general framework for integrating perception capabilities into MLLMs, overcoming the inherent limitations of existing methods.
\end{abstract}

\section{Introduction}

Multimodal large language models (MLLMs)~\cite{bai2023qwen,bai2025qwen2,bai2025qwen3,wang2024qwen2,yang2025qwen3,chen2024internvl,comanici2025gemini,guo2025seed1,lu2024deepseek,zhu2025internvl3} have achieved breakthroughs in bridging vision and language, enabling a wide range of perception tasks that are critical for human-machine interaction, robotics, and autonomous driving. Many previous works~\cite{bai2025qwen2,bai2025qwen3,lai2024lisa,yang2023lisa++,pramanick2024jack,fei2024vitron,tang2025ufo,jiang2025detect,liu2024grounding,liu2025seg,ren2024pixellm,zhang2024psalm,wang2025himtok} integrate 2D or 3D perception capabilities into MLLMs to achieve more general perceptual understanding benefited from the powerful general understanding and reasoning capabilities of MLLMs, yet existing approaches still face fundamental limitations in representation and quantization that hinder their scalability and performance, especially in 3D domains.


As shown in Figure~\ref{fig_intro}, current methods~\cite{bai2025qwen3,bai2025qwen2,guo2025seed1,pramanick2024jack,tang2025ufo,wang2023visionllm,huang2025vision,song2026towards} for MLLM-based perception tasks predominantly rely on text-based representation of regression coordinates. Although these text-based methods can achieve unlimited range and high precision without a minimum quantization unit, they generate an excessive number of tokens, imposing substantial computational overhead in MLLMs. Moreover, these methods make the visual-language learning process more difficult, as the model must learn to identify the structure of floating-point numbers. To alleviate the learning difficulty of continuous coordinate values, a common solution~\cite{jiang2025detect} is to normalize these coordinates into a fixed integer range (e.g., 0 to 1000) and convert continuous spatial information into discrete special tokens that LLMs can process. Although this strategy suffers from two obvious limitations especially from 2D to 3D domains. The quantization of continuous coordinates into a fixed integer range encounters numerical range and precision constraints. In 2D scenes, the finite spatial scope of images partially alleviates these issues, but in 3D scenes, where spatial range is theoretically unbounded and requires precise localization of objects in 3D space~\cite{liu2024lion,hou2024open,brazil2023omni3d}. Due to the limitation of inference efficiency, fixed-range quantization either fails to capture the full spatial information of 3D environments or sacrifices precision due to the limitations of the minimum quantization unit, leading to these methods struggling to achieve full range and high precision of complex 3D scenarios. This raises the question: \textit{Can we achieve 2D and 3D perception representations with unlimited range and high precision using a small number of tokens?}

To address these fundamental limitations, we draw inspiration from Vector Quantized Variational Autoencoders (VQ-VAE)~\cite{kingma2013auto,van2017neural}, a framework that has proven effective in learning compact and discrete representations of continuous data by mapping inputs to a learnable codebook of limited discrete tokens. We argue that compact representations of low-dimensional spaces can be achieved in high-dimensional spaces. Building on this insight, we propose dynamic vector decoding~(DVD) for MLLMs that can seamlessly unify the representation of 2D and 3D perception tasks, effectively overcoming the token inefficiency of text-based coordinate representation and the range and precision constraints of fixed-range quantization. In this paper, our core innovation lies in transforming diverse perception representations (2D boxes, 2D masks, and 3D boxes) into a 1D vector sequence and projecting the sequence into discrete tokens in high-dimensional space, reducing learning difficulty and inference cost for both 2D and 3D tasks. 
Although some previous works~\cite{wu2026liquid,team2024chameleon,jiang2023motiongpt} have explored the integration of VQ-VAE~\cite{van2017neural} into MLLMs, we are the first to introduce a unified high-dimensional codebook for MLLM-based perception, unifying heterogeneous 2D and 3D perceptual representations in high-dimensional space. We also introduce geometric loss to enable the unified codebook to faithfully capture the geometric structures of heterogeneous perception tasks.

Moreover, to learn meaningful discrete representations of these 1D sequences, we design a 1D encoder-decoder architecture, which supports joint training for multiple perception tasks in a unified manner. During training, the encoder maps the 1D vector sequences to a latent space, and the decoder reconstructs the original 1D sequences from this latent representation. Through this process, we learn a unified codebook that maps continuous 1D vector sequences into compact, discrete tokens. This codebook not only enables spatial representation of adaptive quantization for 2D and 3D tasks without fixed range constraints, but also provides inherent compression capabilities, significantly reducing the computational overhead of MLLMs compared to text-based coordinate representation methods. Moreover, a key advantage of our framework is its seamless integration with MLLMs in the decoding process. Because we only need an additional de-tokenizer to decode the tokens belonging to the perception task from the MLLMs output tokens back to their original 2D and 3D representations.

To validate the effectiveness and generality of our proposed method we conduct extensive experiments across 2D and 3D benchmarks for 2D grounding, referring expression segmentation (RES) and 3D grounding tasks. For 3D grounding, we evaluate our method on four widely used datasets: SUN-RGBD~\cite{song2015sun}, KITTI~\cite{geiger2012we}, Hypersim~\cite{roberts2021hypersim}, and nuScenes~\cite{caesar2020nuscenes}, covering diverse indoor and outdoor 3D scenarios. For 2D grounding and RES, we validate our method on the RefCOCO~\cite{lin2014microsoft}, RefCOCO+~\cite{lin2014microsoft}, and RefCOCOg~\cite{mao2016generation} datasets. Experimental results demonstrate that our method outperforms existing state-of-the-art methods in terms of 3D and 2D grounding accuracy, while maintaining competitive performance in 2D referring segmentation and achieving significant reductions in token overhead and inference latency, demonstrating its superiority in both performance and efficiency within a unified framework.

Overall, our contributions are summarized as:
\begin{itemize}
\item We propose a dynamic vector decoding (DVD) that converts 3D boxes, 2D boxes, and 2D masks into 1D vector sequences and learn a unified codebook by projecting these 1D sequences into the high-dimensional space. The proposed method can significantly improve the inference efficiency and provide inherent compression capability.
\item To enable the unified codebook to faithfully capture the geometric structures of heterogeneous perception tasks, we introduce a task-agnostic geometric loss, which supervises the decoded representations in terms of geometry rather than sequence parameterization and provides unified geometric supervision for both 2D and 3D tasks.
\item We validate our method on 2D and 3D perception benchmarks, demonstrating superior performance across 3D grounding task, 2D grounding task, and 2D referring segmentation, while significantly reducing token overhead and inference latency.
\end{itemize}
\section{Related Work}

\paragraph{Multimodal Large Language Models.}
With the rapid development of LLMs in NLP field, MLLMs have become an indispensable part of LLM research, enabling more general intelligent perception and reasoning with the rich visual information. Early MLLMs~\cite{bai2023qwen,bai2025qwen2,bai2025qwen3,wang2024qwen2,yang2025qwen3,guo2025seed1,team2026kimi,chen2024internvl,zhu2025internvl3,wang2025internvl3,team2024gemini,comanici2025gemini} adopt a paradigm of combining vision encoders with pre-trained LLMs via projectors with image-text alignment training. Recent works~\cite{wang2024emu3,yang2025haplovl,diao2024unveiling} have focused on building native MLLMs to unleash stronger general multimodal capabilities. Meanwhile, more and more models such as Qwen3-VL~\cite{bai2025qwen3} and SEED1.5-VL~\cite{guo2025seed1} continuously expand the capabilities of MLLMs by introducing co-training on a large number of different tasks (e.g., Coding, Agent, and Perception), thereby enhancing the versatility of MLLMs. Despite these advancements, most existing MLLMs still face the challenge of how to fully unleash the potential of MLLMs for diverse perception tasks. 

\paragraph{MLLM-based Perception.}
For MLLM-based perception, existing works can mainly be divided into two paradigms. Some works~\cite{lai2024lisa,yang2023lisa++,pramanick2024jack,fei2024vitron,zhang2024psalm,zhang2024omg,pi2024perceptiongpt,chen2023shikra,liu2026drivepi} focus on utilizing additional task-specific decoders. These methods retain the pre-trained MLLM backbone and introduce decoders tailored to specific perceptual tasks (e.g., 2D mask decoding, 3D box regression), leveraging the MLLM’s strong semantic reasoning ability to guide the decoder in generating perceptual outputs. Other works~\cite{bai2025qwen2,bai2025qwen3,tang2025ufo,fu2025wedetect,jiang2025detect,yang2025visual,liu2024grounding,team2026kimi,yang2025visual} directly convert all perception tasks into text format. These methods adopt language to describe perceptual prediction (such as 2D/3D coordinate values), allowing the MLLM to process perception tasks in a unified text-generation manner without additional decoders. However, both paradigms have limitations: task-specific decoders lack generality and cannot simultaneously handle different perception tasks, while text-format methods suffer from the inefficiency of text-based coordinate representation that are more obvious in 3D scenes where spatial dimensions are unbounded and precision requirements are higher.

\paragraph{Perceptual Representation in MLLMs.}
Designing perceptual representation is critical for the performance and efficiency of MLLM-based perception tasks. Most existing methods~\cite{bai2025qwen2,bai2025qwen3,guo2025seed1,wang2025internvl3} rely on text-based representation of regression coordinates. These methods face the difficulty of learning continuous coordinates and need a lot of tokens to process coordinates. To mitigate the difficulty of learning continuous coordinates, some methods normalize coordinates into a fixed integer range (e.g., 0 to 1000) and convert them into discrete textual tokens. For instance, Rex-Omni~\cite{jiang2025detect} propose next-point prediction for different 2D grounding tasks by introducing special tokens for quantized coordinates (0 to 999) in the LLM vocabulary. However, due to the boundless nature of 3D space, this method is not suitable for complex 3D scenes. Additionally, fixed-range quantization has inherent precision constraints, which are more important for accurate localization in 3D scenes.

To address these problems, we propose a dynamic decoding method that unifies 2D and 3D perceptual representations into 1D vector sequences. Our method effectively overcomes limitations of existing  methods by learning a unified codebook for adaptive compression and quantization.
\section{Method}

\subsection{Overview}   
In this paper, we propose a simple and effective MLLM-based perception framework named  dynamic vector decoding (DVD), which can 
unify the representation of 2D and 3D perception tasks and seamlessly integrate into current MLLMs. Specifically, we first convert 2D and 3D perceptual representation into 1D vector sequence. Then we design a 1D encoder-decoder architecture that adaptively maps these different sequences into latent space and generates corresponding discrete tokens by learning unified limited codebook with reconstruction of multiple perception tasks. In the inference, we only need an additional detokenizer to decode the decode the tokens belonging to the perception task from the MLLMs output tokens back to their original 2D/3D representations.

\subsection{Perceptual Representation}
To unify the input format of diverse 2D and 3D perception tasks, we convert all perceptual representation including 2D bounding boxes, 2D masks, and 3D bounding boxes into 1D vector sequences. For 2D boxes, we select left-top and right-down corner points and flatten them $(x_1, y_1, x_2, y_2)$. For 2D masks, we sample fixed points $(x_1, y_1, x_2, y_2,...,x_N,y_N)$ uniformly along the edge of the mask, starting from the top left point in the clockwise direction. We further normalize these points by the width and height of the input image, ensuring that the vector sequence is invariant to image resolution. For 3D boxes, we use the standard structures of 3D boxes $(x,y,z,w,h,l,R)$, where $(x,y,z)$ is the 3D center, $w,h,l$ is the scale, and $R\in \mathbb{R}^{3\times3}$ is the rotation matrix of the 3D box, respectively. Due to the unbounded nature of 3D scenes, we directly use the original numerical values to ensure the precise localization.

\begin{figure*}[t!]
\centering
\includegraphics[width=0.99\linewidth]{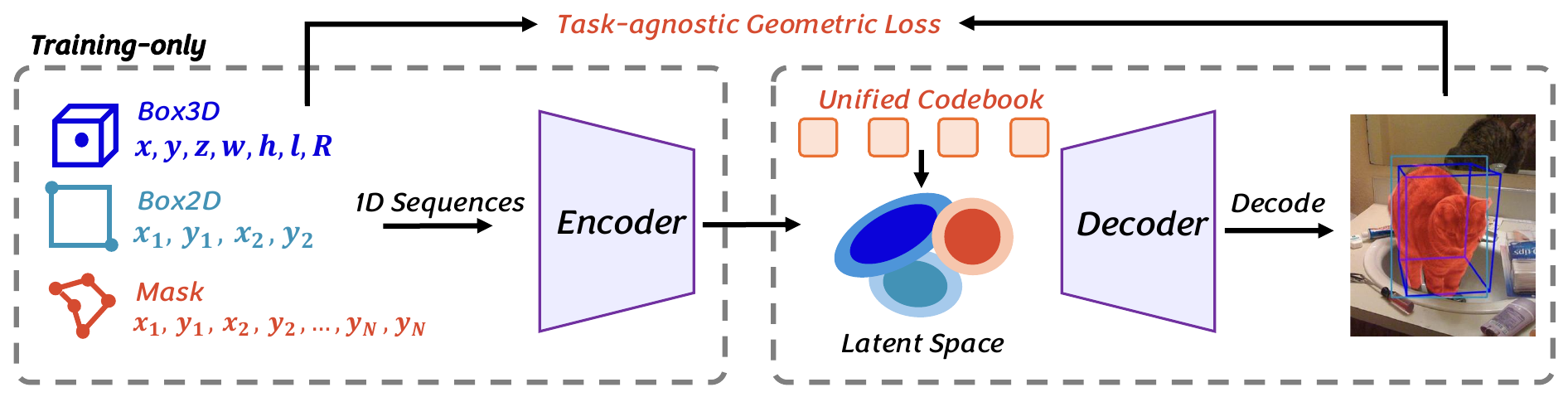}
\caption{
The illustration of the dynamic vector representation learning. We first use the 1D sequence as 2D and 3D perceptual representation. Then, we use an encoder to map these sequences to high-dimensional space to adaptively quantize and compress these 1D information. Moreover, we introduce the task-agnostic geometric loss to improve the heterogeneous perception representation learning capacity. Finally, we use a decoder to recover discrete tokens to original representation.
}
\label{fig_vae}
\end{figure*}

\begin{figure*}[t!]
\centering
\includegraphics[width=0.99\linewidth]{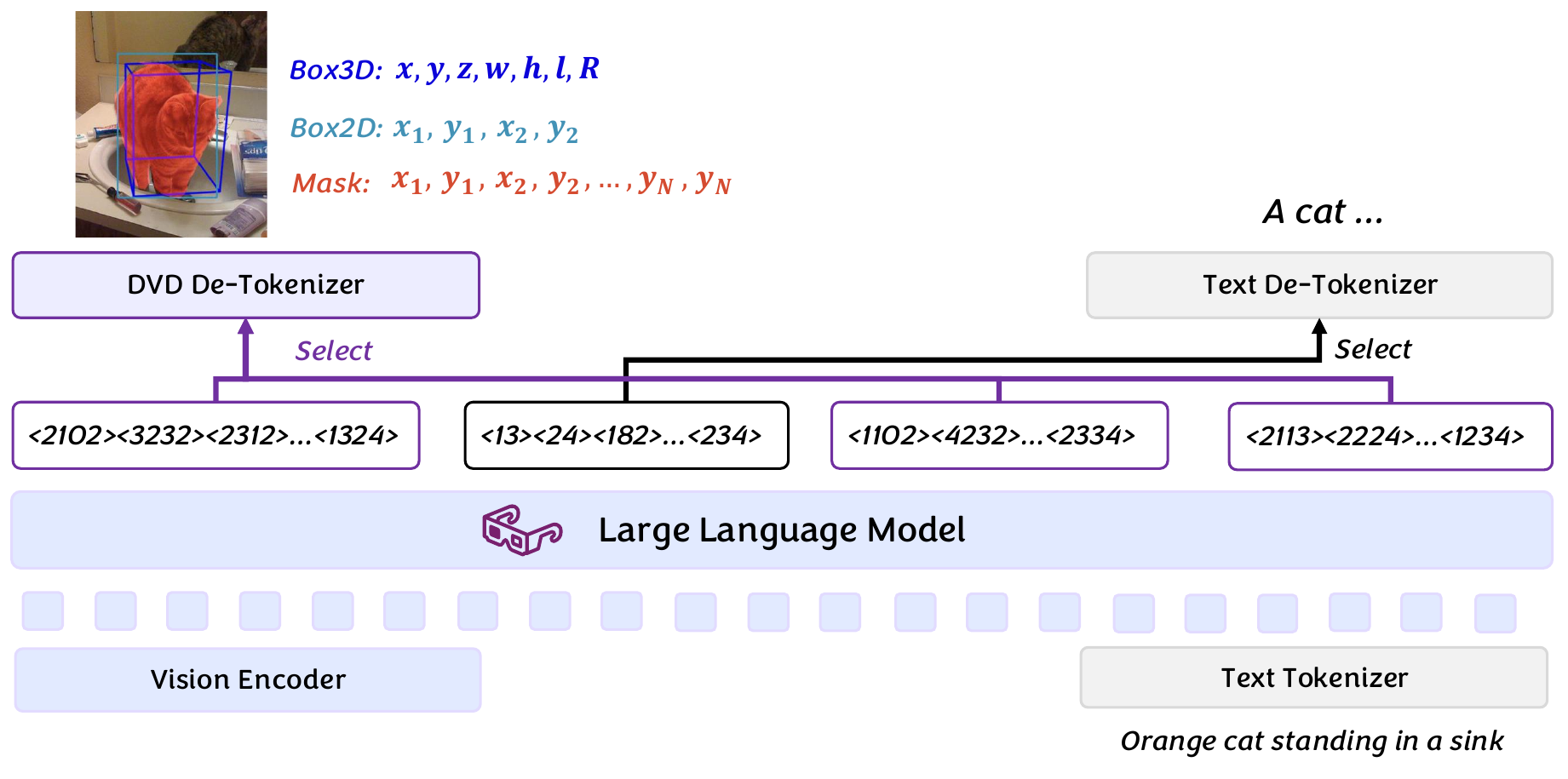}
\vspace{-5pt}
\caption{
Overview of {\ours}. We extend the MLLM vocabulary with learned unified tokens and train the model following standard LLM paradigms without modifications. At inference, we select tokens belonging to perception tasks and use a lightweight de-tokenizer reconstructs the original perceptual outputs from the discrete tokens with minimal overhead, drastically reducing sequence length and improving efficiency.
}
\label{fig_arch}
\end{figure*}

\subsection{Dynamic Vector Representation Learning}

\paragraph{Architecture.}
As shown in Figure~\ref{fig_vae}, the unified vector autoencoder converts 1D vectors to discrete tokens, consisting of a 1D encoder, a learnable codebook, and a 1D decoder. We train the autoencoder end-to-end with reconstruction and multiple perception tasks. Specifically, given the input vector sequences $\mathbf{s}\in \mathbb{R}^{B\times L \times 1}$. For sequences whose length is not divisible by 2, we pad the end of the sequence with zeros. Then we encode input vector sequences into latent space by the 1D encoder, generating latent vectors $\mathbf{z}\in \mathbb{R}^{B\times \frac{L}{K} \times C}$, where $K$ and $C$ are the downsampling factor and feature dimension of vectors, respectively. Next, we compute euclidean distances between flattened $\mathbf{z}$ and embedding vectors to select the closest quantized latent vectors. The symmetric 1D decoder upsamples quantized latent vectors to reconstruct the original 1D sequence. By learning the data distribution after converting the data into a 1D sequence, the vector autoencoder possesses the ability to adapt to different 2D and 3D perception tasks without task-specific modifications. The unified codebook learned from multiple tasks enables the autoencoder to handle variable-length vector sequences from different tasks.

\paragraph{Geometric Training Loss.}
Beyond the standard reconstruction and VQ losses, our unified vector autoencoder requires geometric constraints to faithfully recover the original perceptual representations. This is because the L1 reconstruction loss only measures per-dimension numerical differences of the flattened sequence, which is sensitive to point ordering and parameterization rather than the underlying geometry: two sequences encoding identical shapes with different starting points incur a large penalty, while geometrically distorted outputs may only yield small errors. Meanwhile, the VQ loss constrains the embeddings only in the latent space, providing no guarantee that the decoded outputs preserve geometric structures (e.g., corners, edges, and shapes) in the original representation space. To address these limitations, we introduce a geometric loss based on the bidirectional chamfer distance, which is permutation-invariant and task-agnostic, thus providing a unified geometric supervision for both 2D and 3D perception tasks.

Specifically, our autoencoder is trained with a combined loss function, which includes the reconstruction loss, VQ loss, and geometric loss. The reconstruction loss measures the difference between the original 1D vector sequence $\mathbf{s}$ and the reconstructed sequence $\hat{\mathbf{s}}$, using the L1 loss. The reconstruction loss and VQ loss are formulated as:
\begin{equation}
\mathcal{L}_{\text{L1}} = \mathbb{E}\left\| \mathbf{s} - \hat{\mathbf{s}} \right\|_1
\end{equation}
\begin{equation}
\mathcal{L}_{\text{VQ}} = \mathbb{E}\left\| \mathbf{z}_{q} - \mathbf{z}_{\text{detach}} \right\|_{2}^2 + \beta \cdot \mathbb{E}\left\| \mathbf{z}_{q_{\text{detach}}} - \mathbf{z} \right\|_2^2,
\end{equation}
where $\mathbf{z}$ is the latent vector output by the 1D encoder, $\mathbf{z}_q$ is the quantized latent vector, $\cdot_{\text{detach}}$ denotes the stop-gradient operation, and $\beta$ is a hyperparameter. Beyond the reconstruction loss and VQ loss, we need a geometry guidance to achieve a better learning of inherent geometric characteristic for different perception tasks. Therefore, we apply the geometric loss to enhance geometric learning. For 3D tasks, it computes the average minimum L1 distance between predicted and target 3D box vertices. For 2D tasks, it computes the average minimum L1 distance between predicted and target set of 2D points. The loss is defined as:
\begin{equation}
\mathcal{L}_{\text{geo}} = \frac{1}{m} \sum_{x_i \in X} \min_{y_j \in Y} \left\| x_i - y_j \right\|_1 + \frac{1}{n} \sum_{y_j \in Y} \min_{x_i \in X} \left\| y_j - x_i \right\|_1,
\end{equation}
where $X$ and $Y$ are the predicted and target point set, respectively. The $m$ and $n$ denote the number of predicted and target points. The total loss is defined as:
\begin{equation}
\mathcal{L}_{\text{total}} = \lambda_1 \cdot \mathcal{L}_{\text{L1}} + \lambda_2 \cdot \mathcal{L}_{\text{VQ}} + \lambda_3 \cdot \mathcal{L}_{\text{geo}},
\end{equation}
where $\lambda_1$, $\lambda_2$, $\lambda_3$ is the weight of the reconstruction loss, VQ loss, and geometric loss.

\subsection{Dynamic Vector Decoding}
To integrate the unified vector autoencoder with MLLMs, we first extend the vocabulary of the MLLM by adding $M$ new tokens, corresponding to the $M$ code vectors in the codebook. During the training of the MLLM with our autoencoder, we use the standard training paradigm as the same as MLLMs. This integration is lightweight and does not require modifying the MLLM backbone, ensuring compatibility with various pre-trained MLLMs. In the inference, we use the decoder of our vector autoencoder as the de-tokenizer for converting the discrete tokens output by the MLLM back into the 1D vector sequence, which is then converted into the original 2D/3D perceptual representation. Specifically, we first use the de-tokenizer to map each discrete token belonging to perception tasks output by the MLLM to its corresponding code vector in the pre-trained codebook. Then the de-tokenizer uses the pre-trained 1D decoder to reconstruct the 1D vector sequence from the code vector. Notably, the dynamic vector decoding process introduces minimal computational overhead compared to text-based methods. Since each perceptual representation is mapped to a small number of discrete tokens and the sequence length fed into the MLLM is significantly reduced, improving inference efficiency. Additionally, the de-tokenizer is a lightweight module (reusing the pre-trained decoder), avoiding the need for complex task-specific decoders and maintaining the generality of the framework.
\section{Experiments}

\subsection{Datasets}
For 3D grounding, we use four benchmarks covering indoor and outdoor scenarios: SUN-RGBD~\cite{song2015sun}, Hypersim~\cite{roberts2021hypersim}, KITTI~\cite{geiger2012we}, and nuScenes~\cite{caesar2020nuscenes}. SUN-RGBD~\cite{song2015sun} is a 3D indoor dataset with RGB-D images and 3D bounding box annotations. Hypersim~\cite{roberts2021hypersim} is a photorealistic synthetic indoor dataset, providing 3D bounding boxes and semantic labels for generalization validation. KITTI~\cite{geiger2012we} is an outdoor autonomous driving dataset, annotated with 3D bounding boxes for traffic objects. nuScenes~\cite{caesar2020nuscenes} is a comprehensive outdoor benchmark with 1000 scenes and high-precision 3D bounding boxes. For 2D grounding and referring segmentation tasks, we use RefCOCO~\cite{lin2014microsoft}, RefCOCO+~\cite{lin2014microsoft}, and RefCOCOg~\cite{mao2016generation} as benchmarks. RefCOCO~\cite{lin2014microsoft} is built on the COCO~\cite{lin2014microsoft} dataset with additional referring expressions. RefCOCO+~\cite{lin2014microsoft} excludes location-reliant expressions. RefCOCOg~\cite{mao2016generation} further features complex expressions.

\subsection{Implementation Details}
To train our unified vector autoencoder, we collect approximately 1.98M 3D boxes, 2.31M 2D boxes, and 1.52M 2D masks in a 1:1:1 data ratio during training. The downsampling scale $K$ is set to 2. The codebook size and latent hidden channel is set to 4096 and 16, respectively. The $\lambda_1$, $\lambda_2$, and $\lambda_3$ are set to 1. We train the autoencoder with the learning rate of $10^{-3}$ for 24 epochs. For MLLM, we use the Qwen3-VL~\cite{bai2025qwen3}, consisting of SigLIP2~\cite{tschannen2025siglip}, vision encoder MLP-based projector, and Qwen3 LLM~\cite{yang2025qwen3}. as our pretrained model. For training, we use a batch size of 64 with AdamW~\cite{loshchilov2017decoupled} optimizer and a data packing strategy to accelerate the training on. The base learning rate is to $5\times 10^{-5}$ and the learning rate of vision encoder is to $5\times 10^{-6}$. We fully finetune the MLLM for 264K iterations with 2.0M training QAs curated from mixed 2D and 3D perception annotations of 2D and 3D datasets. All experiments are conducted on 8 NVIDIA A100 GPUs.

\begin{table*}[t!]
\caption{Performance comparison of 2D and 3D perception.
}
\small
\setlength{\tabcolsep}{2pt}
\resizebox{1.0\linewidth}{!}{
\centering
\begin{tabular}{l|c|c|c|c|c|c|c|c|c}
\toprule
\multicolumn{1}{c|}{\multirow{2}{*}{Method}} & \multicolumn{3}{c|}{\emph{3D Grounding}} & \multicolumn{3}{c|}{\emph{2D Grounding}} & \multicolumn{3}{c}{\emph{2D RES}}\\
& SUN-RGBD & Hypersim & nuScenes & RefCOCO & RefCOCO+ & RefCOCOg & RefCOCO & RefCOCO+ & RefCOCOg \\
\midrule
GroundingDINO~\cite{liu2024grounding} & \textbf{--} & \textbf{--} & \textbf{--} & 90.6 & 88.2 & 86.1 & \textbf{--} & \textbf{--} & \textbf{--} \\
VistaLLM-7B~\cite{pramanick2024jack} & \textbf{--} & \textbf{--} & \textbf{--} & 88.1 & 82.9 & 83.6 & 74.5 & 69.1 & 69.0 \\
LISA-7B~\cite{lai2024lisa} & \textbf{--} & \textbf{--} & \textbf{--} & \textbf{--} & \textbf{--} & \textbf{--} & 74.9 & 65.1 & 67.9 \\
Text4Seg~\cite{lan2025text4seg} & \textbf{--} & \textbf{--} & \textbf{--} & 88.3 & 83.5 & 82.4 & 74.7 & 68.5 & 70.7 \\
Qwen2.5-VL-3B~\cite{bai2025qwen2} & \textbf{--} & \textbf{--} & \textbf{--} & 89.1 & 82.4 & 85.2 & \textbf{--} & \textbf{--} & \textbf{--} \\
Qwen2.5-VL-7B~\cite{bai2025qwen2} & \textbf{--} & \textbf{--} & \textbf{--} & 90.0 & 84.2 & 87.2 & \textbf{--} & \textbf{--} & \textbf{--} \\
Rex-Omni~\cite{jiang2025detect} & \textbf{--} & \textbf{--} & \textbf{--} & \textbf{--} & \textbf{--} & 86.6  & \textbf{--} & \textbf{--} & \textbf{--} \\
Seed1.5-VL~\cite{guo2025seed1} & 33.5 & \textbf{--} & \textbf{--} & \textbf{--} & \textbf{--} & \textbf{--} & \textbf{--} & \textbf{--} & \textbf{--}\\
Qwen3-VL-2B~\cite{bai2025qwen3} & 33.8 & 12.0 & \textbf{--} & 89.2 & 80.4 & 84.0 & \textbf{--} & \textbf{--} & \textbf{--} \\
Qwen3-VL-8B~\cite{bai2025qwen3} & 36.2 & 12.7 & \textbf{--} & 91.6 & 86.1 & 87.7 & \textbf{--} & \textbf{--} & \textbf{--} \\
VST-3B~\cite{yang2025visual} & 37.3 & \textbf{--} & \textbf{--} & \textbf{--} & \textbf{--} & \textbf{--} & \textbf{--} & \textbf{--} & \textbf{--} \\
\midrule
\rowcolor{blue!4}
{\ours}-2B & 40.0 & 15.7 & 19.8 & \textbf{93.5} & \textbf{87.7} & \textbf{89.9} & 69.9 & 64.2 & 70.4 \\
\rowcolor{blue!4}
{\ours}-8B & \textbf{40.3} & \textbf{18.4} & \textbf{24.4} & 93.0 & \textbf{87.7} & 88.7 & \textbf{75.4} & \textbf{70.1} & \textbf{73.4} \\
\bottomrule
\end{tabular}
 }
\label{tab:main}
\vspace{-10pt}
\end{table*}

\subsection{Overall Performance}
We follow previous MLLM-based perception methods and evaluate our method on the 3D grounding, 2D grounding, and 2D referring segmentation tasks. For 3D grounding, we use the \textit{AP3D@15} defined by Omni3D~\cite{brazil2023omni3d} as the metric. For 2D grounding and referring segmentation, we use \textit{Precision@0.5} and \textit{cIoU} as metrics. Note that we use the same trained model to validate the performance of all tasks. 

\paragraph{MLLM-based Perception Comparison.}
Table~\ref{tab:main} compares {\ours} with representative MLLM-based perception methods on all three tasks. Benefiting from dynamic vector decoding, which overcomes the limitations of text-based coordinate representation and fixed-range quantization, {\ours}-2B surpasses SOTA methods on 3D grounding with 40.0\%, 15.7\%, and 19.8\% AP3D on SUN-RGBD~\cite{song2015sun}, Hypersim~\cite{roberts2021hypersim}, and nuScenes~\cite{caesar2020nuscenes}, and achieves 89.9\% Precision and 70.4\% cIoU on RefCOCOg val~\cite{kazemzadeh2014referitgame}. With a larger backbone, {\ours}-8B further reaches 40.3\%, 18.4\%, and 24.4\% AP3D on the three 3D datasets, and 88.7\% Precision and 73.4\% cIoU on RefCOCOg val. These results demonstrate the effectiveness of dynamic vector decoding in enhancing the perception capabilities of MLLMs.

\begin{table*}[t!]
\caption{Performance comparison of 3D grounding on both indoor and outdoor datasets.
}
\small
\setlength{\tabcolsep}{17pt}
\resizebox{1.0\linewidth}{!}{
\centering
\begin{tabular}{l|c|c|c|c|c}
\toprule
Method & SUN-RGBD & Hypersim & ARKitScenes & KITTI & nuScenes \\
\midrule
Gemini 2.0 Pro~\cite{team2024gemini} & 32.5 & \textbf{--} & \textbf{--} & \textbf{--} & \textbf{--} \\
Gemini 2.5 Pro~\cite{comanici2025gemini} & 29.7 & \textbf{--} & \textbf{--} & \textbf{--} & \textbf{--} \\
Seed1.5-VL~\cite{guo2025seed1} & 33.5 & \textbf{--} & \textbf{--} & \textbf{--} & \textbf{--} \\
Qwen3-VL-2B~\cite{bai2025qwen3} & 33.8 & 12.0 & \textbf{--} & \textbf{--} & \textbf{--} \\
Qwen3-VL-8B~\cite{bai2025qwen3} & 36.2 & 12.7 & \textbf{--} & \textbf{--} & \textbf{--} \\
VST-3B~\cite{yang2025visual} & 37.3 & \textbf{--} & 51.7 & \textbf{--} & \textbf{--} \\
\midrule
\rowcolor{blue!4}
{\ours}-2B & 40.0 & {15.7} & \textbf{62.4} & \textbf{31.4} & {19.8} \\
\rowcolor{blue!4}
{\ours}-8B & \textbf{40.3} & \textbf{18.4} & {62.1} & {28.7} & \textbf{24.4} \\
\bottomrule
\end{tabular}
 }
\label{tab:3D}
\vspace{-10pt}
\end{table*}

\paragraph{3D Grounding.}
We further compare with MLLM-based 3D perception methods on SUN-RGBD~\cite{song2015sun}, Hypersim~\cite{roberts2021hypersim}, ARKitScenes~\cite{baruch2021arkitscenes}, KITTI~\cite{geiger2012we}, and nuScenes~\cite{caesar2020nuscenes} (Table~\ref{tab:3D}), covering both indoor and outdoor scenes, with annotations and metrics from Omni3D~\cite{brazil2023omni3d}. {\ours}-2B outperforms VST-3B~\cite{yang2025visual} by 2.7\% and 10.7\% AP3D on SUN-RGBD and ARKitScenes, while {\ours}-8B achieves 40.3\% and 24.4\% AP3D on SUN-RGBD and nuScenes.
\begin{table*}[t!]
\caption{Performance comparison of 2D grounding on the RefCOCO series.
}
\small
\setlength{\tabcolsep}{12pt}
\resizebox{1.0\linewidth}{!}{
\centering
\begin{tabular}{l|c|c|c|c|c|c|c|c}
\toprule
\multicolumn{1}{c|}{\multirow{2}{*}{Method}} & \multicolumn{3}{c|}{\emph{RefCOCO}} & \multicolumn{3}{c|}{\emph{RefCOCO+}} & \multicolumn{2}{c}{\emph{RefCOCOg}}\\
& val & testA & testB & val & testA & testB & val & test \\
\midrule
GroundingDINO~\cite{liu2024grounding} & 90.6 & 93.2 & 88.2 & 88.2 & 89.0 & 75.9 & 86.1 & 87.0\\
Qwen2.5-VL-7B~\cite{bai2025qwen2} & 90.0 & 92.5 & 85.4 & 84.2 & 89.1 & 76.9 & 87.2 & 87.2\\
Rex-Omni~\cite{jiang2025detect} & \textbf{--} & \textbf{--} & \textbf{--} & \textbf{--} & \textbf{--} & \textbf{--} & 86.6 & 86.8 \\
Seed1.5-VL~\cite{guo2025seed1} & \textbf{--} & \textbf{--} & \textbf{--} & \textbf{--} & \textbf{--} & \textbf{--} & 84.7 & 85.2 \\
InternVL3.5-8B~\cite{wang2025internvl3} & 92.4 & 94.7 & 88.7 & 87.9 & 92.4 & 82.4 & {89.6} & {89.4} \\
UFO-8B~\cite{tang2025ufo} & 91.8 & 94.3 & 87.5 & 86.9 & 91.3 & 80.6 & 87.9 & 88.6 \\
Qwen3-VL-8B~\cite{bai2025qwen3} & 91.6 & 93.1 & 88.7 & 86.1 & 91.1 & 80.0 & 87.7 & 88.4\\
\midrule
\rowcolor{blue!4}
{\ours}-2B & \textbf{93.5} & \textbf{95.5} & \textbf{90.7} & \textbf{87.7} & \textbf{92.9} & {82.4} & \textbf{89.9} & \textbf{90.0} \\
\rowcolor{blue!4}
{\ours}-8B & {93.0} & \textbf{95.2} & {89.7} & \textbf{87.7} & \textbf{92.9} & \textbf{82.7} & 88.7 & 89.8 \\
\bottomrule
\end{tabular}
 }
\vspace{-15pt}
\label{tab:2D}
\end{table*}

\paragraph{2D Grounding.}
As shown in Table~\ref{tab:2D}, we compare {\ours} with representative grounding methods (e.g., GroundingDINO~\cite{liu2024grounding}, Qwen series~\cite{bai2025qwen2,bai2025qwen3}, Rex-Omni~\cite{jiang2025detect}). Although designed to unify 2D and 3D perception tasks, {\ours} still outperforms advanced 2D grounding methods: {\ours}-2B achieves 90.0\% Precision on RefCOCOg test, surpassing Rex-Omni~\cite{jiang2025detect} by 3.2\%.

\begin{table*}[t!]
\caption{Performance comparison of 2D referring expression segmentation on the RefCOCO series.
}
\small
\setlength{\tabcolsep}{12pt}
\resizebox{1.0\linewidth}{!}{
\centering
\begin{tabular}{l|c|c|c|c|c|c|c|c}
\toprule
\multicolumn{1}{c|}{\multirow{2}{*}{Method}} & \multicolumn{3}{c|}{\emph{RefCOCO}} & \multicolumn{3}{c|}{\emph{RefCOCO+}} & \multicolumn{2}{c}{\emph{RefCOCOg}}\\
& val & testA & testB & val & testA & testB & val & test \\
\midrule
VLT~\cite{ding2021vision} & 67.5 & 70.5 & 65.2 & 56.3 & 61.0 & 50.1 & 55.0 & 57.7 \\
LAVT~\cite{yang2022lavt} & 72.7 & 75.8 & 68.8 & 62.1 & 68.4 & 55.1 & 61.2 & 62.1 \\ 
VistaLLM-7B~\cite{pramanick2024jack} & 74.5 & 76.0 & {72.7} & {69.1} & 73.7 & {64.0} & 69.0 & 70.9 \\
LISA-7B~\cite{lai2024lisa} & 74.9 & 79.1 & 72.3 & 65.1 & 70.8 & 58.1 & 67.9 & 70.6 \\
M$^{2}$SA~\cite{jang2025mmr} & 74.0 & 76.8 & 69.7 & 63.1 & 67.2 & 56.1 & 67.0 & 68.3 \\
PixelLM-7B~\cite{ren2024pixellm} & 73.0 & 76.5 & 68.2 & 66.3 & 71.7 & 58.3 & 69.3 & 70.5 \\
Text4Seg~\cite{lan2025text4seg} & 74.7 & 77.4 & 71.6 & 68.5 & 73.6 & 62.9 & 70.7 & 71.6 \\
PerceptionGPT-7B~\cite{pi2024perceptiongpt} & {75.1} & \textbf{78.6} & 71.7 & 68.5 & {73.9} & 61.3 & {70.3} & {71.7} \\
\midrule
\rowcolor{blue!4}
{\ours}-2B & 69.9 & 70.7 & 69.7 & 64.2 & 68.2 & 61.8 & 70.4 & 71.0  \\
\rowcolor{blue!4}
{\ours}-8B & \textbf{75.4} & 76.9 & \textbf{74.6} & \textbf{70.1} & \textbf{74.7} & \textbf{65.9} & \textbf{73.4} & \textbf{76.6}  \\
\bottomrule
\end{tabular}
 }
\vspace{-10pt}
\label{tab:res}
\end{table*}

\paragraph{Referring Expression Segmentation.}
Table~\ref{tab:res} reports 2D referring expression segmentation (RES) results. Despite having far fewer parameters and being a unified model, {\ours}-2B remains competitive with VistaLLM-7B~\cite{pramanick2024jack} on RefCOCOg test~\cite{kazemzadeh2014referitgame} with lower latency thanks to the reduced token overhead. With a larger backbone, {\ours}-8B further achieves 73.4\% and 76.6\% cIoU on RefCOCOg val and test. This shows that dynamic vector decoding generalizes beyond box-based grounding to fine-grained segmentation.


\begin{figure*}[t!]
\centering
\includegraphics[width=0.98\linewidth]{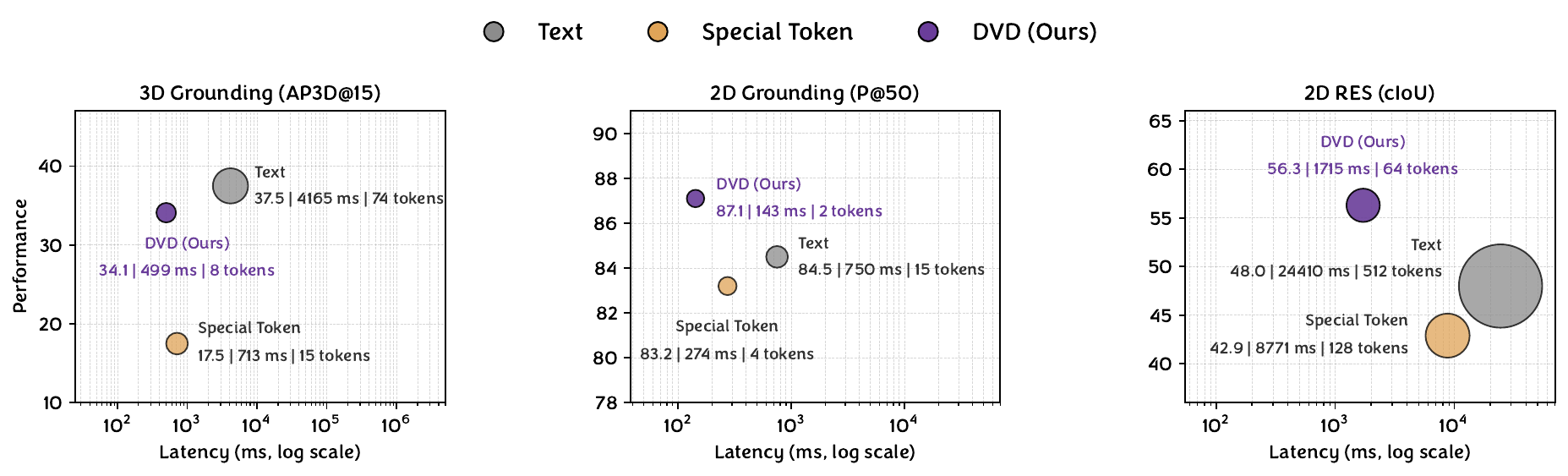}
\vspace{-5pt}
\caption{
Computation cost and performance comparison of different representations. Bubble area indicates the token count generated for per object of the representation.
}
\label{fig_cost}
\vspace{-15pt}
\end{figure*}


\paragraph{Vector Reconstruction Performance.}
The quality of 1D vector reconstruction directly affects the performance of subsequent perception tasks. Therefore, we evaluate the reconstruction performance of our unified vector autoencoder by using perception metrics between the ground truth and the decoded sequences from the quantized tokens. Our autoencoder achieves negligible reconstruction error (90.8\% AP3D, 100.0\% Precision, and 84.8\% cIoU for 3D grounding, 2D grounding, and 2D RES task), demonstrating its ability to preserve spatial information.


\subsection{Ablation Study}
In this section, we adopt ablation study to illustrate the effectiveness of our method. For convenience, we train our model with 30\% of the full training iterations for faster validation, and all ablation experiments are conducted on the {\ours}-2B. More ablations and visualization results are illustrated in the appendix.

\paragraph{Comparison of different representations.}
To demonstrate the efficiency of {\ours}, we compare the average number of tokens generated for per object, latency, and performance of different representations in Figure~\ref{fig_cost}, using identical training settings and computation resources for a fair comparison. For the special token representation, we follow the coordinate tokenization of Rex-Omni~\cite{jiang2025detect} and extend the vocabulary with 4000 special tokens shared across all coordinates: 2D box and mask coordinates are normalized to $[0,1]$ by image size, while 3D centers, sizes, and rotation entries are scaled into $[-1,1]$ and linearly mapped to $[0,1]$, then uniformly quantized into 4000 bins.

We report the average tokens generated per object and the inference latency on SUN-RGBD~\cite{song2015sun} and RefCOCOg~\cite{kazemzadeh2014referitgame}. 
Text-based coordinates generate excessive tokens (512 and 74 for mask and 3D box), incurring high latency. Since text preserves the raw coordinate values with unlimited range and precision, it retains a slight accuracy advantage on 3D grounding 37.5\% vs.\ 34.1\% AP3D). However, this marginal gain comes with 9$\times$ more tokens and 8.3$\times$ higher latency (4165 ms vs.\ 499 ms per object). Special tokens reduce the cost but suffer severe degradation (17.5\% AP3D) due to limited range and precision. In contrast, {\ours} matches text-level precision on 2D tasks, largely closes the 3D gap, and achieves significant 3D gains the special-token representation, delivering the best accuracy efficiency trade-off.

\begin{table*}[t!]
\centering 
\begin{minipage}{0.48\textwidth}
\centering
\caption{The effect of codebook size.}
\small
\setlength{\tabcolsep}{3pt}
\resizebox{\linewidth}{!}{
\begin{tabular}{c|c|c|c}
\toprule
\multirow{2}{*}{Codebook} & \emph{3D Grounding} & \emph{2D Grounding} & \emph{2D RES}\\
& AP3D@15 \emph{avg.} & P@50 \emph{avg.} & cIoU \emph{avg.}\\
\midrule
2048 & 88.9 & 84.5 & 81.6 \\
\rowcolor{blue!4}
4096 & \textbf{90.8} & \textbf{100.0} & \textbf{84.8} \\
8192 & 85.1 & 99.6 & 76.8 \\
\bottomrule
\end{tabular}
}
\label{tab:code}
\end{minipage}
\hfill
\begin{minipage}{0.50\textwidth}
\centering
\caption{The effect of geometric loss.}
\small
\setlength{\tabcolsep}{3pt}
\resizebox{\linewidth}{!}{
\begin{tabular}{c|c|c|c}
\toprule
\multirow{2}{*}{Geo. Loss} & \emph{3D Grounding} & \emph{2D Grounding} & \emph{2D RES}\\
& AP3D@15 \emph{avg.} & P@50 \emph{avg.} & cIoU \emph{avg.}\\
\midrule
 & 80.7 & 99.1 & 74.8 \\
\rowcolor{blue!4}
\checkmark & \textbf{90.8} & \textbf{100.0} & \textbf{84.8} \\
\bottomrule
\end{tabular}
}
\label{tab:geo}
\end{minipage}
\vspace{-10pt}
\end{table*}



\noindent \textbf{Codebook Size.} Table~\ref{tab:code} illustrates the effect of the codebook size. A small codebook (2048) limits the representational capacity of the discrete latent space, failing to cover diverse geometric patterns (88.9\% AP3D and 84.5\% Precision). On the other hand, an overly large codebook (8192) increases the difficulty of quantization and codebook learning, leading to under-trained entries and degraded reconstruction (85.1\% AP3D and 76.8\% cIoU). The codebook size of 4096 achieves the best performance across all three tasks and is thus adopted as our default setting.



\noindent \textbf{Geometric Loss.} Table~\ref{tab:geo} validates the effect of the geometric loss. Incorporating geometric guidance consistently improves the reconstruction performance across all tasks, with particularly notable gains on 3D grounding (+10.1\% AP3D) and 2D RES (+10.0\% cIoU). This indicates that geometric supervision plays a critical role when fine-grained spatial fidelity is required. For example, depth estimation in 3D scenes and boundary delineation in segmentation are both highly sensitive to small coordinate deviations. Therefore, the vanilla reconstruction objective alone fails to penalize. In contrast, the improvement on 2D grounding is relatively marginal (+0.9\% Precision), since coarse box localization in 2D images is less sensitive to minor geometric errors and its performance is already near saturation. These observations confirm that the geometric loss is not merely a regularizer but an essential component for learning a unified vector representation. By explicitly constraining the decoded vectors to preserve the underlying spatial structure, it enables a single codebook to serve both 2D and 3D box-based grounding and mask-based segmentation without task-specific designs. 

\begin{table*}[t!]
\centering 
\begin{minipage}{0.48\textwidth}
\centering
\caption{The effect of downsampling scale.}
\small
\setlength{\tabcolsep}{5pt}
\resizebox{\linewidth}{!}{
\begin{tabular}{c|c|c|c}
\toprule
\multirow{2}{*}{Scale} & \emph{3D Grounding} & \emph{2D Grounding} & \emph{2D RES}\\
& AP3D@15 \emph{avg.} & P@50 \emph{avg.} & cIoU \emph{avg.}\\
\midrule
1 & 89.3 & \textbf{100.0} & \textbf{92.7} \\
\rowcolor{blue!4}
2 & \textbf{90.8} & \textbf{100.0} & 84.8 \\
4 & 83.7 & 31.5 & 60.4 \\
\bottomrule
\end{tabular}
}
\label{tab:scale}
\end{minipage}
\hfill
\begin{minipage}{0.50\textwidth}
\centering
\caption{The effect of latent hidden size.}
\small
\setlength{\tabcolsep}{5pt}
\resizebox{\linewidth}{!}{
\begin{tabular}{c|c|c|c}
\toprule
\multirow{2}{*}{z} & \emph{3D Grounding} & \emph{2D Grounding} & \emph{2D RES}\\
& AP3D@15 \emph{avg.} & P@50 \emph{avg.} & cIoU \emph{avg.}\\
\midrule
8 & 83.0 & \textbf{100.0} & 83.7 \\
\rowcolor{blue!4}
16 & \textbf{90.8} & \textbf{100.0} & 84.8 \\
32 & 88.4 & \textbf{100.0} & \textbf{85.3} \\
\bottomrule
\end{tabular}
}
\label{tab:latent}
\end{minipage}
\end{table*}

\begin{table*}[t!]
\caption{Reconstruction Performance comparison of multi-task training.
}
\small
\setlength{\tabcolsep}{12pt}
\resizebox{1.0\linewidth}{!}{
\centering
\begin{tabular}{c|c|c|c|c|c}
\toprule
3D Grounding & 2D Grounding & 2D RES & AP3D@15 \emph{avg.} & P@50 \emph{avg.} & cIoU \emph{avg.} \\
\midrule
\checkmark & & & 84.0 & \textbf{--} & \textbf{--} \\
& \checkmark & & \textbf{--} & \textbf{100.0} & \textbf{--} \\
& & \checkmark & \textbf{--} & \textbf{--} & {80.3} \\
\rowcolor{blue!4}
\checkmark & \checkmark & \checkmark & \textbf{90.8} & \textbf{100.0} & \textbf{84.8} \\
\bottomrule
\end{tabular}
 }
\label{tab:multitask}
\vspace{-10pt}
\end{table*}

We further investigate the effect of different settings of the DVD autoencoder on the vector reconstruction performance, including the downsampling scale $K$, latent hidden size $z$, and multi-task learning in the vector representation learning.

\noindent \textbf{Downsampling Scale.} As shown in Table~\ref{tab:scale}, we ablate the effect of the downsampling scale $K$. Compared with the uncompressed representation ($K{=}1$), setting $K{=}2$ reduces the token count by half while maintaining competitive performance (90.8\% AP3D and 100.0\% Precision for 3D and 2D grounding, respectively), and even slightly improves the 3D grounding accuracy. However, an excessively large scale ($K{=}4$) leads to severe information loss during compression, resulting in significant performance degradation, especially on the 2D grounding task (31.5\% Precision). These results demonstrate that $K{=}2$ achieves the best trade-off between reconstruction precision and token cost, verifying that our learned codebook retains spatial semantics while reducing sequence length.

\noindent \textbf{Latent Hidden Size.} As shown in Table~\ref{tab:latent}, we ablate the latent hidden size $z$. A narrow latent bottleneck ($z{=}8$) constrains the encoded spatial information, causing a notable drop on the 3D grounding task (83.0\% AP3D). In contrast, enlarging $z$ to 32 brings no consistent gains and even slightly degrades the 3D reconstruction quality (88.4\% AP3D), as a higher-dimensional latent space is harder to quantize effectively. Setting $z{=}16$ provides sufficient capacity with stable quantization, achieving the best overall performance.

\noindent \textbf{Multi-task Training in DVD.}
Our proposed DVD supports 2D and 3D perception tasks with unified 1D vector representation. We validate the effect of multi-task training by comparing the performance of the autoencoder trained on single task and multi-task training in Table~\ref{tab:multitask}. Experimental results show that joint multi-task training improve performance on both 3D grounding task and 2D RES task. Moreover, the performance of 2D grounding remains stable without obvious degradation. These results demonstrate the learning capability for dynamic vector representation of our method without sacrificing the performance of individual tasks.

\begin{figure*}[t!]
\centering
\includegraphics[width=0.99\linewidth]{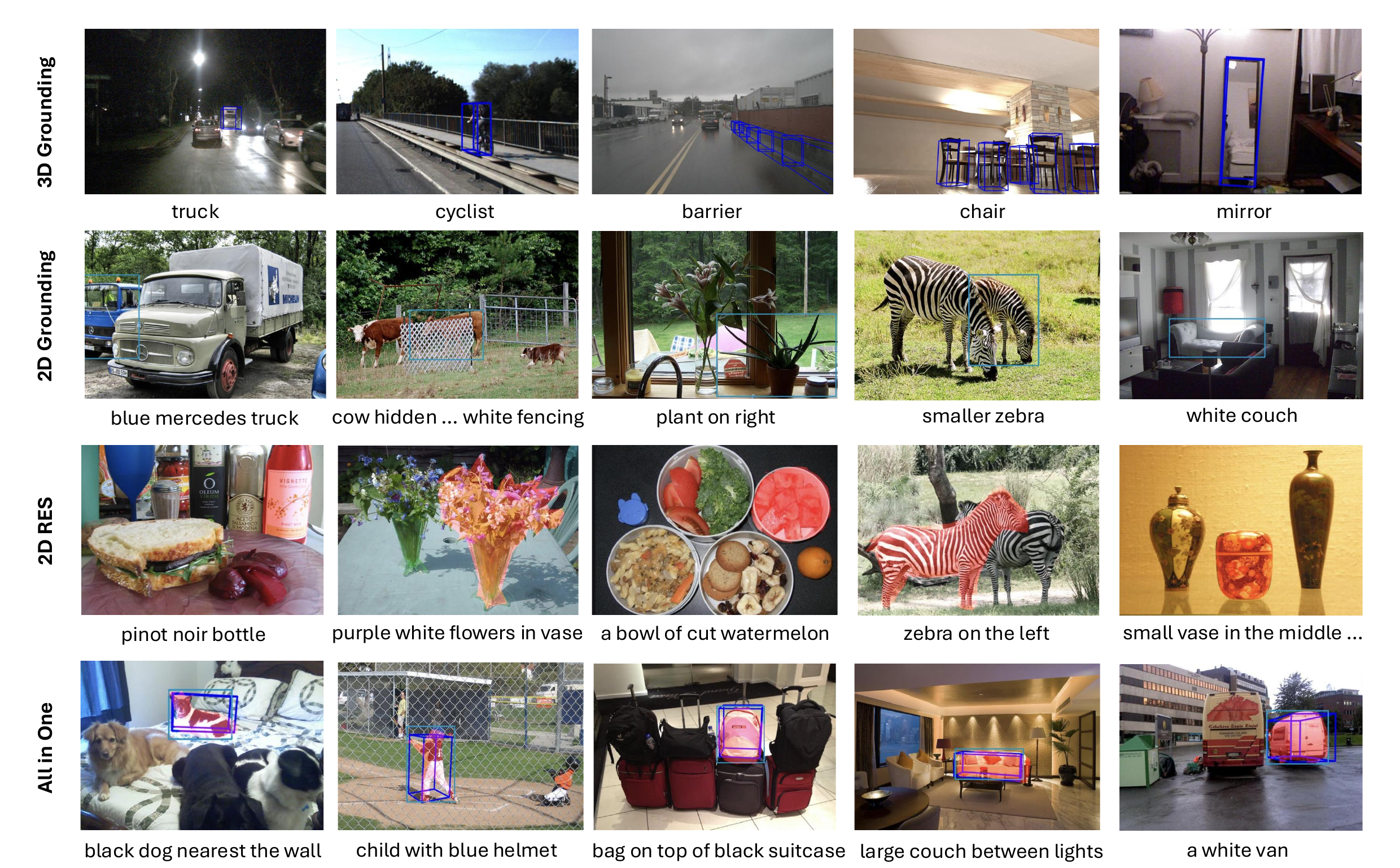}
\caption{
The qualitative visualization of {\ours} on the 2D and 3D perception tasks. The first row is the results of 3D grounding task. The second row is the result of 2D grounding task. The third row is the results of 2D RES task. The fourth row is the results of unified perception.
}
\label{fig_vis1}
\end{figure*}

\subsection{Visualization.}

We provide qualitative visualization results to illustrate the perception capability of {\ours}. As shown in Figure~\ref{fig_vis1}, we visualize the qualitative results of {\ours} on the 3D grounding, 2D grounding, and 2D RES tasks. As shown in the first row, {\ours} can achieve satisfactory results even in dark and complex 3D scenes. In the second and third row, {\ours} achieves precise 2D localization and segmentation capabilities in different objects and different scenes. In the fourth row, {\ours} can simultaneously achieves 2D and 3D perception tasks, demonstrating the superior perception capability of {\ours}.

\section{Conclusion}
Existing MLLM-based perception methods struggle with excessive token overhead from text-based coordinate representation and precision and range constraints from fixed-range quantization, especially in 3D domains. To address the key limitation, we propose DVD that unifies 2D and 3D perception task representations for efficient integration with MLLMs. By transforming heterogeneous perceptual outputs into 1D vector sequences and mapping them to compact discrete tokens, DVD eliminates the excessive token overhead and the precision limitations. Extensive experiments on 2D and 3D benchmarks show that DVD achieves superior performance while significantly reducing token count and inference latency, offering an efficient and general framework for equipping MLLMs with perception capabilities.

{\small
\bibliographystyle{plain}
\bibliography{ref}
}

\newpage
\appendix
\section{Technical Appendices}
Our technical appendices provide additional information, including implementation details of the proposed unified vector autoencoder,
dataset details, more ablation studies, and visualization results.

\section{Implementation Details}
In this section, we provide the architecture details of our unified autoencoder. The 1D encoder is designed with a multi-stage downsampling architecture, stacked with 1D resnet blocks and 1D attention blocks. It takes 1D vector sequences as input, and extracts feature representations through successive layers and downsampling operations. A $1\times1$ convolutional layer is appended to the encoder output to project the latent features into the embedding dimension of the codebook.

The vector quantization module adopts the VectorQuantizer~\cite{van2017neural}, which maintains a fixed-size learnable discrete codebook. This module computes the Euclidean distance between the projected latent features and each codebook embedding, assigns each feature to the nearest codebook entry, and adopts straight-through gradient estimation to ensure gradient consistency. The codebook serves as the core of the unified representation, mapping continuous 1D vector sequences into compact discrete tokens that can be efficiently processed by MLLMs.
Symmetric to the encoder, the 1D decoder adopts a multi-stage upsampling architecture, also composed of 1D resnet blocks and 1D attention blocks. A $1\times1$ convolutional layer is placed at the decoder input to project the quantized discrete tokens back to the latent dimension. The decoder then reconstructs the original 1D vector sequences from the quantized tokens through successive layers and upsampling operations. The latent dimension and codebook is set to 16 and 4096, respectively.

\section{Dataset Details}


\begin{table}[h!]
\centering
\caption{Dataset Details.}
\label{tab:datasets}
\setlength{\tabcolsep}{6pt}
\small
\begin{tabularx}{\linewidth}{c|>{\raggedright\arraybackslash}X|c}
\toprule
Task & Datasets & QAs \\
\midrule
\rowcolor{blue!4}
3D Grounding 
& SUN-RGBD~\cite{song2015sun}, Hypersim~\cite{roberts2021hypersim}, ARKitScenes~\cite{baruch2021arkitscenes}, Objectron~\cite{ahmadyan2021objectron}, KITTI~\cite{geiger2012we}, nuScenes~\cite{caesar2020nuscenes} 
& 489K \\
2D Grounding 
& RefCOCO~\cite{lin2014microsoft}, RefCOCO+~\cite{lin2014microsoft}, RefCOCOg~\cite{mao2016generation} \newline
SUN-RGBD~\cite{song2015sun}, Hypersim~\cite{roberts2021hypersim}, ARKitScenes~\cite{baruch2021arkitscenes}, Objectron~\cite{ahmadyan2021objectron}, KITTI~\cite{geiger2012we}, nuScenes~\cite{caesar2020nuscenes}
& 810K \\
\rowcolor{blue!4}
2D RES 
& RefCOCO~\cite{lin2014microsoft}, RefCOCO+~\cite{lin2014microsoft}, RefCOCOg~\cite{mao2016generation}, RefCLEF~\cite{kazemzadeh2014referitgame}, ReasonSeg~\cite{lai2024lisa}, COCO~\cite{lin2014microsoft}
& 774K \\
\bottomrule
\end{tabularx}
\end{table}

To comprehensively evaluate the effectiveness, generality, and robustness of {\ours} across 2D and 3D perception tasks, we conduct experiments on a diverse set of benchmark datasets covering indoor and outdoor scenes. These datasets are carefully curated from existing public datasets, covering the core scenarios of 3D grounding, 2D grounding, and 2D RES. Detailed statistics of the datasets are summarized in Table~\ref{tab:datasets}. For 3D grounding, we select six representative datasets to cover both indoor and outdoor 3D scenarios. SUN-RGBD~\cite{song2015sun} and ARKitScenes~\cite{baruch2021arkitscenes} focus on indoor environments with diverse household objects. Hypersim~\cite{roberts2021hypersim} provides photorealistic synthetic scenes. Objectron~\cite{ahmadyan2021objectron} focuses on single real-world 3D object. KITTI~\cite{geiger2012we} and nuScenes~\cite{caesar2020nuscenes} are outdoor autonomous driving datasets with complex traffic scenes and dynamic objects. These datasets provide 489K 3D grounding QAs.
For 2D grounding, we combine two types of datasets: three standard 2D referring datasets (RefCOCO~\cite{lin2014microsoft}, RefCOCO+~\cite{lin2014microsoft}, RefCOCOg~\cite{mao2016generation}) with rich instructions and diverse 2D scenes, and the 2D projections of the six 3D datasets mentioned above. This combination ensures a total of 810K 2D grounding QAs, allowing {\ours} to achieve better generalization across 2D and 3D scenarios.
For 2D RES, we use six datasets (RefCOCO~\cite{lin2014microsoft}, RefCOCO+~\cite{lin2014microsoft}, RefCOCOg~\cite{mao2016generation}, RefCLEF~\cite{kazemzadeh2014referitgame}, ReasonSeg~\cite{lai2024lisa}, COCO~\cite{lin2014microsoft}) that cover diverse objects, scenes, and instruction granularity. With 774K QAs, these datasets enable {\ours} to achieve fine-grained perceptual capability.


\section{More Ablation Studies}

\begin{table*}[t!]
\caption{Computation cost and performance comparison of different representations.
}
\small
\setlength{\tabcolsep}{5pt}
\resizebox{1.0\linewidth}{!}{
\centering
\begin{tabular}{l|c|c|c|c|c|c|c|c|c}
\toprule
\multirow{2}{*}{Method} & \multicolumn{3}{c|}{\emph{3D Grounding}} & \multicolumn{3}{c|}{\emph{2D Grounding}} & \multicolumn{3}{c}{\emph{2D RES}}\\
& tokens & latency (ms) & AP3D@15 & tokens & latency (ms) & P@50 & tokens & latency (ms) & cIoU \\
\midrule
Text & 74 & 4165 & \textbf{37.5} & 15 & 750 & 84.5 & 512 & 24410 & 48.0 \\
Special Token & 15 & 713 & 17.5 & 4 & 274 & 83.2 & 128 & 8771 & 42.9 \\
\midrule
\rowcolor{blue!4}
{\ours} & \textbf{8} & \textbf{499} & 34.1 & \textbf{2} & \textbf{143} & \textbf{87.1} & \textbf{64} & \textbf{1715} & \textbf{56.3} \\
\bottomrule
\end{tabular}
 }
\label{tab:cost}
\vspace{-10pt}
\end{table*}

\subsection{Comparison of different representations}

As shown in Table~\ref{tab:cost}, we report the detailed results of average number of tokens generated per object and the average inference latency across 3D grounding, 2D grounding, and 2D referring segmentation tasks on SUN-RGBD~\cite{song2015sun} and RefCOCOg~\cite{kazemzadeh2014referitgame} datasets. Existing MLLM-based methods rely on text-based coordinate representation, which generates excessive tokens (512 and 74 tokens for mask and 3D bounding box, respectively) and thus increases inference latency. In contrast, our {\ours} leverages the unified codebook to map 1D vector sequences into compact discrete tokens, achieving significant reductions in token count, leading to much faster inference speed. Although special token achieves obvious computation cost reduction, this representation brings significant performance degradation because it struggle with the limited range and precision. Compared to special token, {\ours} achieves a significant performance improvement in the 3D scene, illustrating the superiority of our method in precise localization capabilities. These results demonstrate that our dynamic vector decoding method not only maintains strong performance on perception tasks but also achieves remarkable efficiency.

\begin{table*}[t!]
\caption{The Codebook Utilization.}
\setlength{\tabcolsep}{40pt}
\resizebox{1.0\linewidth}{!}{
\centering
\begin{tabular}{l|c|c}
\toprule
Task & Tokens & Active Tokens  \\
\midrule
2D boxes & 0.11M & 2957/4096 (72.19\%) \\
3D boxes & 2.27M & 4095/4096 (99.98\%)\\
2D masks & 4.26M & 3940/4096 (96.19\%)\\
\midrule
Overall & 6.65M & 4095/4096 (99.98\%) \\
\bottomrule
\end{tabular}
 }
\label{tab:utilization}
\end{table*}

\begin{figure*}[t!]
\centering
\includegraphics[width=0.98\linewidth]{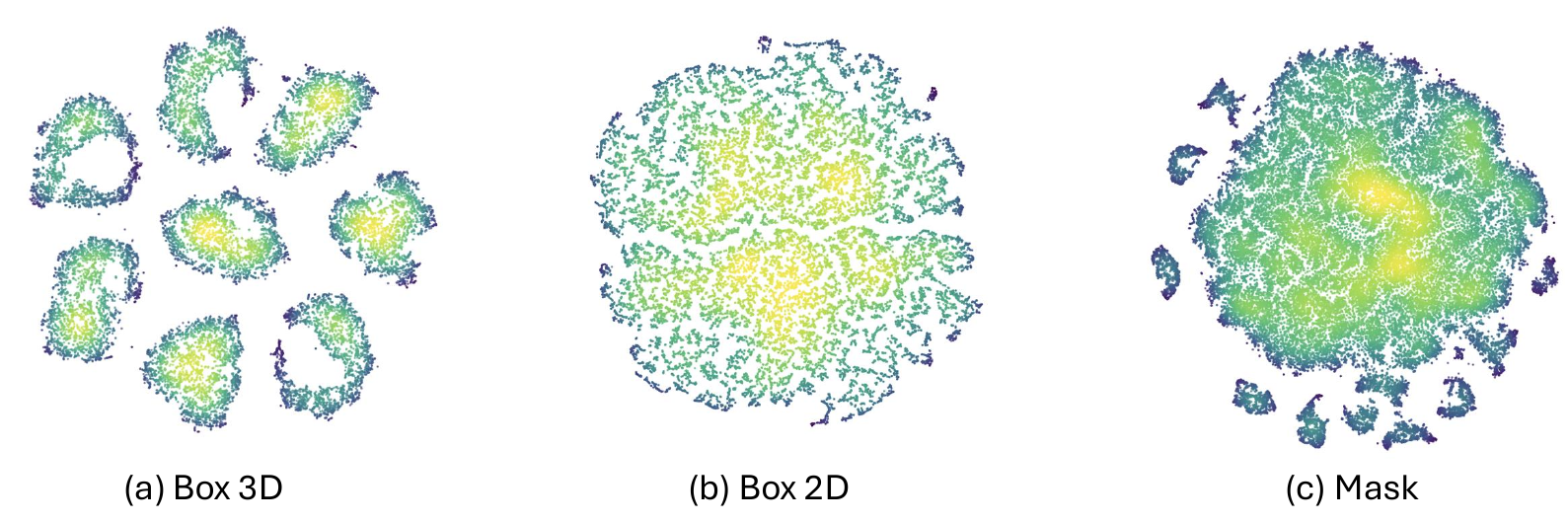}
\caption{
The visualization of latent space in {\ours} with t-SNE. (a) The vector latent space of 3D boxes. (b) The vector latent space of 3D boxes. (c) The vector latent space of 2D masks.
}
\label{fig_tsne}
\end{figure*}

\begin{figure*}[t!]
\centering
\includegraphics[width=0.4\linewidth]{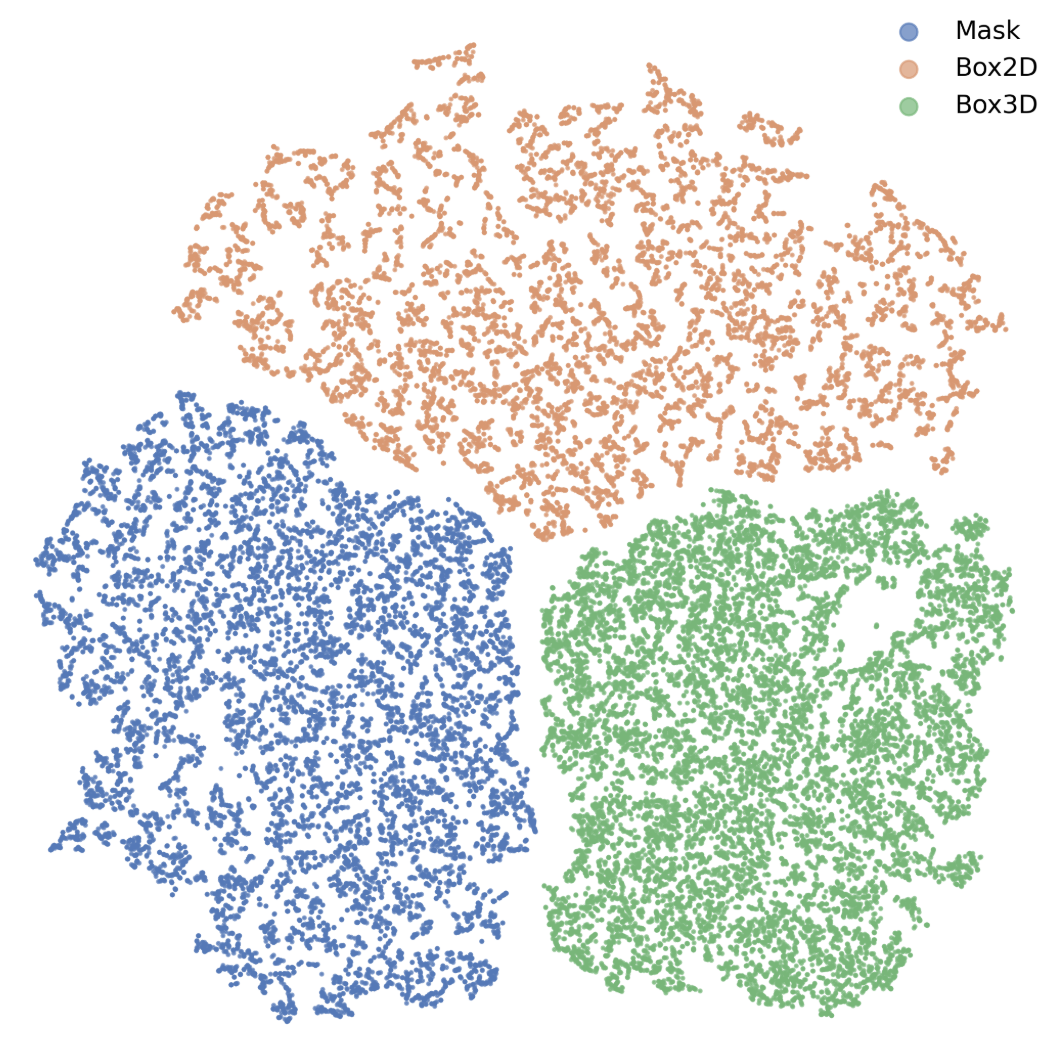}
\caption{
The visualization of latent space in {\ours} with t-SNE for unifying 2D and 3D perception tasks. There are there obvious clusters, illustrating the effectiveness of our dynamic vector representation learning process.
}
\label{fig_tsne2}
\end{figure*}

\subsection{Vector Latent Distribution}
We provide visualization of the learned vector latent space of discrete tokens in {\ours}. As shown in Figure~\ref{fig_tsne}, the results show that there are obvious clusters for distributions of different perception tasks. For example, we use 8 tokens to represent one 3D box and there are 8 clusters in the vector latent space. These results demonstrate that {\ours} effectively learn high-quality latent representations for 1D vector representation. 

We also visualize the distribution in the high-dimensional space of different perception vector representations to demonstrate the effectiveness of {\ours} in unifying 2D and 3D perception tasks. As shown in Figure~\ref{fig_tsne2}, {\ours} produces there latent clusters in high-dimensional space for 2D and 3D perception tasks. 

\subsection{The Codebook Utilization}
We additionally provide the codebook utilization on the validation set to investigate the learning quality of latent space. As shown in Table~\ref{tab:utilization}, the codebook is healthy and near-saturated (99.98\% active codes overall) with no sign of codebook collapse. Consistent with the task-aware clusters in Figure~\ref{fig_tsne} and Figure~\ref{fig_tsne2}, each task occupies its own subset of codes. The simplest representation 2D box (only 2 tokens per box) achieves activate fewer codes (72.2\%), while fine-grained masks spread over more codes with the highest usage uniformity.


\section{Visualization}

\begin{figure*}[t!]
\centering
\includegraphics[width=0.85\linewidth]{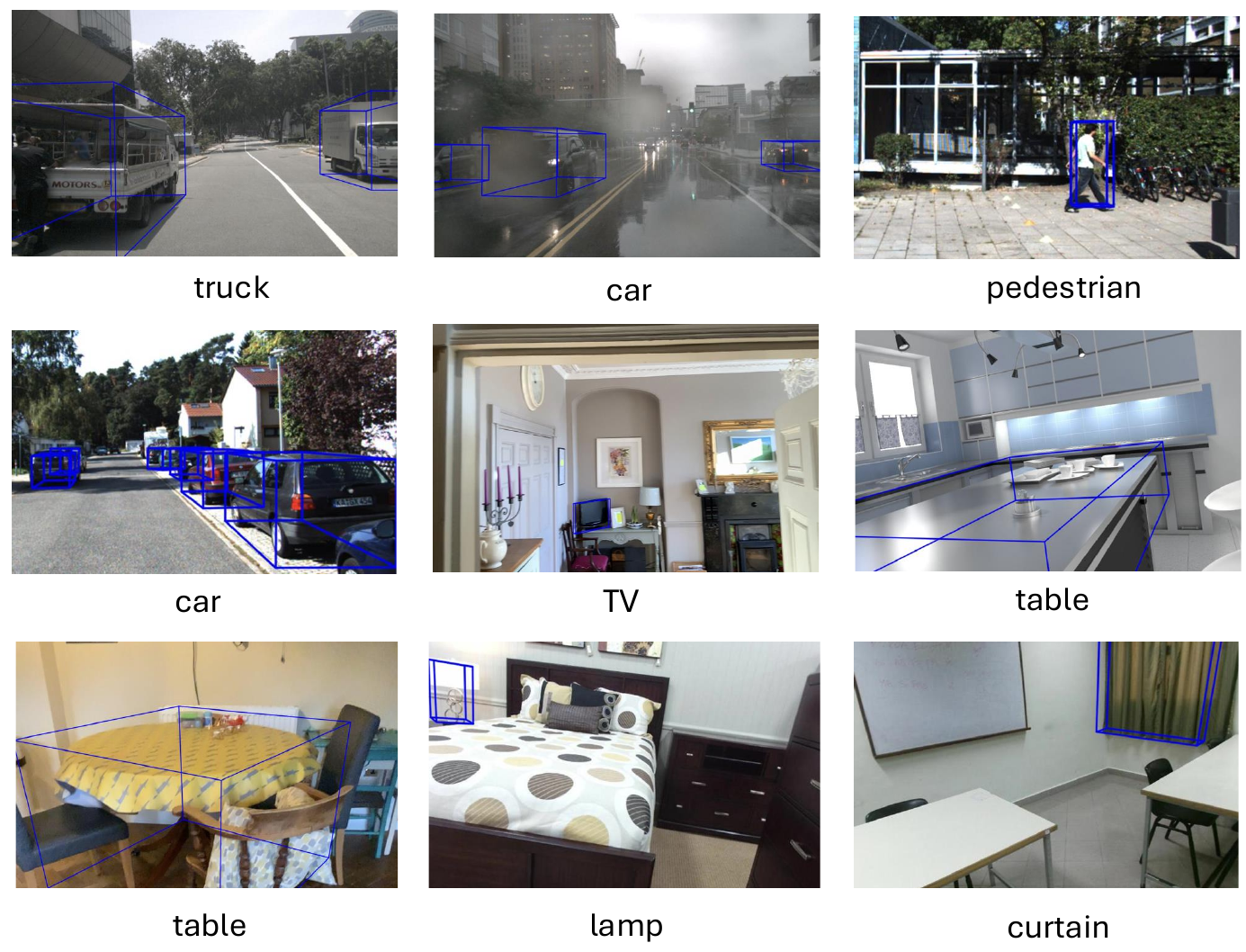}
\caption{
The qualitative visualization of {\ours} on the 3D grounding task.
}
\label{fig_vis2}
\end{figure*}

We further present more detailed visualization results to fully demonstrate the superior perception capabilities of {\ours} across both 2D and 3D perception tasks. As shown in Figure~\ref{fig_vis2}, we present qualitative visualization results of {\ours} on the 3D grounding task, covering both indoor and outdoor scenarios. Notably, even under challenging conditions (e.g., rainy weather, cluttered 3D scenes with dense objects) {\ours} consistently achieves precise 3D localization.

\begin{figure*}[t!]
\centering
\includegraphics[width=0.85\linewidth]{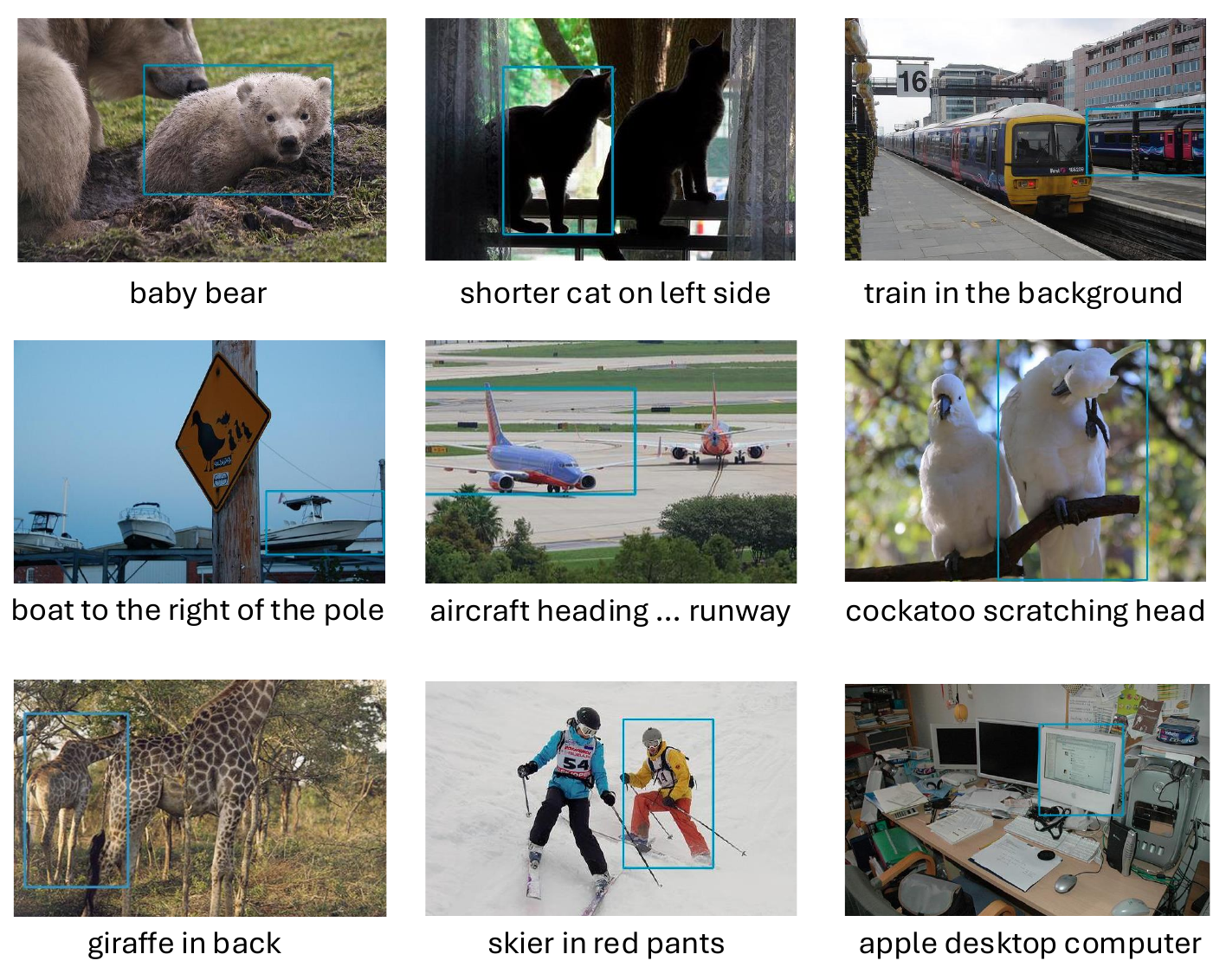}
\caption{
The qualitative visualization of {\ours} on the 2D grounding task.
}
\label{fig_vis3}
\end{figure*}

In Figure~\ref{fig_vis3}, we showcase the results of {\ours} on the 2D grounding task. {\ours} accurately follows instructions and outputs tightly aligned 2D bounding boxes, without misalignment, highlighting its exceptional ability to visual localization. 

\begin{figure*}[t!]
\centering
\includegraphics[width=0.85\linewidth]{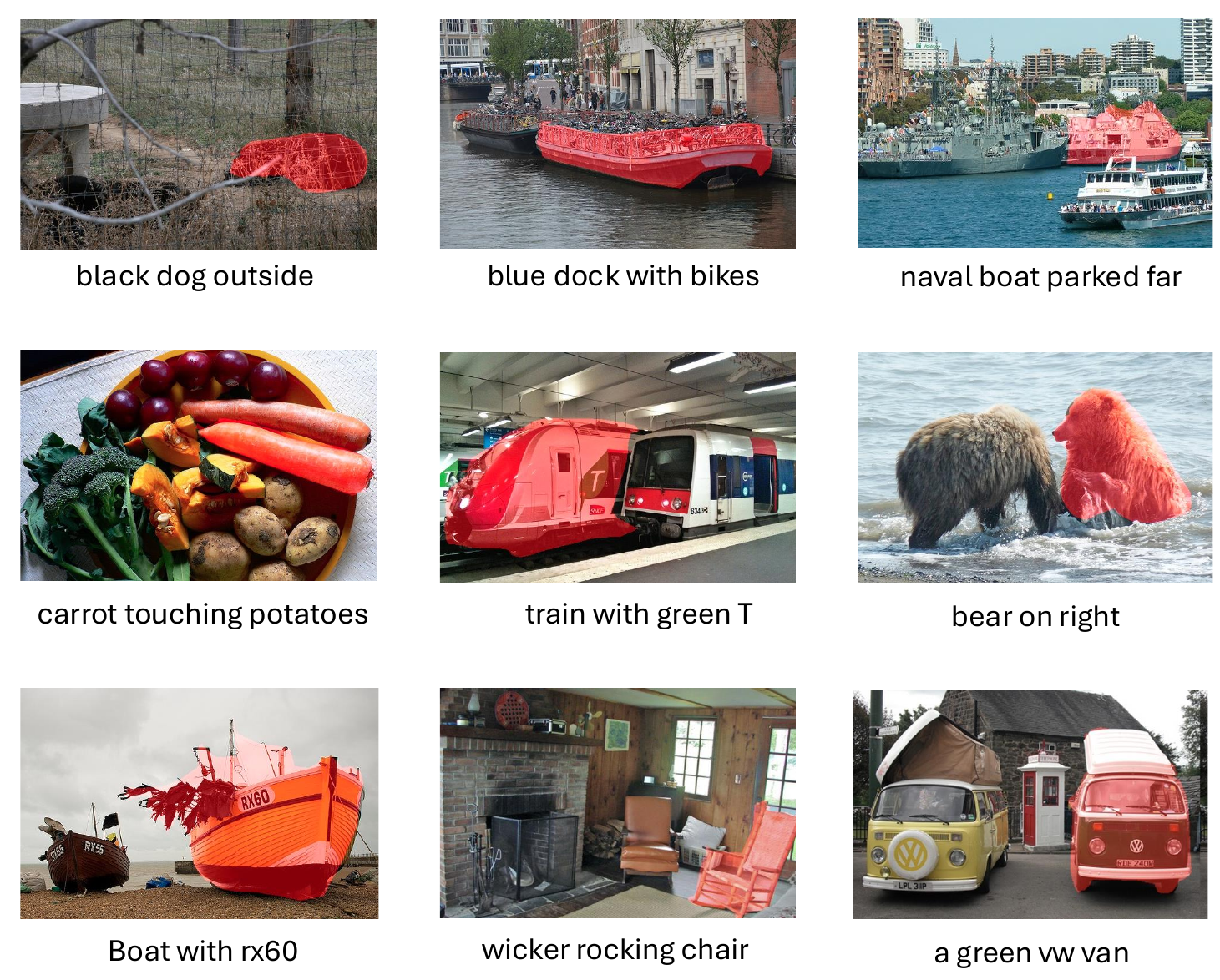}
\caption{
The qualitative visualization of {\ours} on the 2D RES task.
}
\label{fig_vis4}
\end{figure*}

Additionally, Figure~\ref{fig_vis4} includes visualizations of segmentation masks generated by {\ours} for diverse objects under various instruction prompts. {\ours} precisely segments target objects of arbitrary shapes, even in cases with ambiguous boundaries or overlapping objects, further verifying its fine-grained perceptual capability.

\section{Limitations and Future Works}
\subsection{Limitations}
Despite the advantages of our Dynamic Vector Decoding (DVD) framework in unifying 2D and 3D perception for MLLMs, it has some limitations. The reconstruction performance of our dynamic vector representation learning can still be improved to achieve better performance even with a larger downsampling factor. Moreover, the DVD de-tokenizer for codebook learning, while effective, is designed as a separate component from the MLLM. This decoupling may lead to suboptimal alignment between the learned discrete tokens and the semantic understanding of MLLM.
\subsection{Future Works}
Our method can support more representations of more perception tasks. Future work will further optimize the dynamic vector representation learning process,  extend the codebook to diverse tasks, and explore end-to-end training of the encoder-decoder and MLLM.

\section{Broader Impacts}
Our DVD framework provides a new generalizable paradigm for efficient MLLM-based perception. On the one hand, it benefits autonomous driving and robotics by reducing overhead and unifying 2D and 3D perception tasks. On the other hand, it reduces the difficulty of deploying large models on the edge by significantly reducing computational overhead, and lowers the research threshold to promote innovation for MLLMs. 

However, due to its powerful perception capabilities, our research could be misused in closed-circuit television surveillance systems and affect personal privacy.


\end{document}